\documentclass[10pt,conference]{IEEEtran}
 \usepackage{amssymb}
 \usepackage{amsthm}
 \usepackage{amsmath}
 \usepackage{dirtree}
\usepackage{graphicx} % Required for inserting images
\usepackage{lipsum}
\usepackage{tikz}
\usetikzlibrary{arrows.meta}
\usetikzlibrary{arrows,shapes,snakes,automata,backgrounds,petri,positioning,decorations.markings}
\usepackage{quantikz}
\usepackage{cases}
\usepackage{subcaption}
\usepackage{cite}
\usepackage{hyperref}
\usepackage{mathabx}
\usepackage{multirow, booktabs}
\usepackage{graphicx,subcaption,ragged2e}
\usepackage{algorithm, algpseudocode}

\newcommand{\reals}{\mathbb{R}}
\newcommand{\clues}{{\small\textsf{Clues}}}

\newtheorem{proposition}{Proposition}

\title{Constrained Hierarchical Clustering\\ via Graph Coarsening and Optimal Cuts}

\author{%%
\IEEEauthorblockN{Eliabelle Mauduit$^{1, 2}$, Andrea Simonetto$^1$}
\vspace*{2mm}
\IEEEauthorblockA{$^1$ Unité de Mathématiques Appliquées, ENSTA Paris, Institut Polytechnique de Paris, 91120 Palaiseau, France}
\IEEEauthorblockA{$^2$ Greenworking, 110 rue de Picpus, 75012 Paris, France}}

\begin{document}

\maketitle

% 4-6 pages plus references. Abstract due April 10, Paper April 17. 

\begin{abstract}
   Motivated by extracting and summarizing relevant information in short sentence settings, such as satisfaction questionnaires, hotel reviews, and X/Twitter, we study the problem of clustering words in a hierarchical fashion. In particular, we focus on the problem of clustering with horizontal and vertical structural constraints. Horizontal constraints are typically cannot-link and must-link among words, while vertical constraints are precedence constraints among cluster levels. 

   We overcome state-of-the-art bottlenecks by formulating the problem in two steps: first, as a soft-constrained regularized least-squares which guides the result of a sequential graph coarsening algorithm towards the horizontal feasible set. Then, flat clusters are extracted from the resulting hierarchical tree by computing optimal cut heights based on the available constraints.
   
   We show that the resulting approach compares very well with respect to existing algorithms and is computationally light. %The constraints are easy to add on-the-fly, thereby making the algorithm adapted to user interaction. 
\end{abstract}


\section{Introduction}

In modern society, extracting, classifying, and summarizing information from text is becoming of paramount importance. This is fuelled by the large amount of content that is available on social media platforms, and on the Internet in general. In this paper, we are particularly interested in extracting topics and clustering words in short sentences. This text modality is typical in satisfaction questionnaires, X/Twitter, and most of the online posts. Hierarchical clustering is widely used for genomics data~\cite{eiser1998clus}, social network analysis \cite{jure2014mini}, bioinformatics \cite{ibai2015nove}, text classification \cite{stein2000comp}, and financial markets \cite{tumm2010corr}.

Specifically, we focus here on clustering words from short sentences in a hierarchical fashion and adding structural constraints. For most real-life applications, users might have prior information which is not contained in the input features, hence the interest to specify structural constraints. Since the constraints come from contextual information about the data, our problem falls in the category of semi-supervised clustering. Constrained versions of clustering and hierarchical clustering have appeared in the past~\cite{basu2004active, davidson2005agglomerative, bade2006personalized, kawale2013constrained, bair2013semi, florence2017constrained, carlsson2018hierarchical, chatziafratis2018hierarchical, huang2018hierarchical, liu2019clustering} and the area stays very active \cite{bakkelund2022order, cai2023review}. Here, we propose an efficient algorithm to combine both \emph{horizontal constraints}, meaning must-link and cannot-link constraints between words, and \emph{vertical constraints}, meaning precedence constraints among different layers, or levels, of the hierarchy. The aforementioned features make our algorithm novel and unique in its genre. 

We overcome some of the bottlenecks encountered in the literature by considering a two-steps approach.  {\bf First}, we develop a soft-constrained version of both vertical and horizontal constraints added as a regularization term to a graph coarsening algorithm. This enables to take the prior knowledge into account while respecting the initial structure of the data space, and makes it possible to deal with conflicting sets of constraints 
\cite{chatziafratis2018hierarchical}. The result of this step is a dendrogram, which is a tree that iteratively splits a data set into smaller subsets until each subset consists of only one element. {\bf Second}, the obtained dendrogram is then cut at different levels in order to get hierarchical flat clusters that meet as many constraints as possible. The key of our two-steps approach is to automatically get the desired flat clusters, as we will discuss shortly. 

Another particular feature of our approach is that we will consider vertical and horizontal constraints per layer. Since in hierarchical clustering, all the data points will eventually belong to the same set, the constraints have to change or evolve layer per layer. Two data points may have a cannot link at a lower layer, but then they can be merged at a higher level. We call this algorithmic feature: layer-based constraints.
\smallskip

{\bf Related work.} Hierarchical and semi-supervised clustering has been studied extensively in the past~\cite{cai2023review}. Most existing hierarchical clustering are based on linkage distance methods, and the main differential is in the way prior information is handled. The standard way is to add pairwise constraints \cite{wags2000clus, davidson2005agglomerative, yang2022semi} which are basically horizontal must-links and/or cannot-links. While such constraints are a good starting point, they hierarchical layer-based feature is missing. One way to add this is to consider triple-wise constraints instead as in~\cite{zhen2011semi}. However, this approach has the drawback that it  does not handle cannot-links and does not guarantee that the merging heights of the different triplets will define clear levels.

On the constraint side, when hard constraints are imposed~\cite{wags2000clus}, lower complexities can be reached ($\mathcal{O}(n\log(n))$ where $n$ is the number of data points) to the potential detriment of the geometry of the initial data space. However, hard constraints might lead to errors when the set of constraints is not well defined or infeasible. Thus, even though using soft constraints increases algorithmic complexity ($\mathcal{O}(n^2)$ for \cite{zhen2011semi, yang2022semi} and $\mathcal{O}(n^3)$ for the alternative in \cite{zhen2011semi}), they might be preferable to work with to avoid infeasibility.

Another traditional method to perform clustering is to use graph-based reduction algorithms. In such a framework, the space is represented as a graph, and among the different strategies we can cite single-graph \cite{wei2012spec,SEMERTZIDIS2015616} and multiple graphs clustering \cite{LIU201919}. Contrary to single graph approach which works on one graph only and its Laplacian, multiple graphs techniques cluster the data points based on the knowledge provided by several graphs. This can enhance the performances of the clustering process compared to single-graph but it also increases the memory cost. 

All the methods introduced above are either non hierarchical or do no result in flat clusters. To our knowledge, very few methods exist to perform automatic cluster extraction from a hierarchical structure. In their article \cite{jorg2003auto}, J{\"o}rg and co-authors introduce a method to deduce a hierarchical partitioning of the data based on the reachability plot resulting from the hierarchical clustering. Their algorithm is based on the identification of clear ``dents" or ``valleys" in the reachability plot and perform very well on examples where clusters are clearly separated in the initial data structure. Nevertheless, in our case, the data structure does not consist of clean and well defined (sub)-clusters and the algorithm from \cite{jorg2003auto} cannot be directly applied. 

\smallskip
{\bf Contributions.} In this paper, we introduce a new semi-supervised algorithm to perform hierarchical flat clustering. Our algorithms emerge as the natural extensions of A.~Loukas's local-variation multi-level graph coarsening \cite{loukas2019graph} and of the regular bottom-up clustering \cite{mullner2011modern}. These methods rely on the Laplacian and on the distance matrix of the graph associated to the dataset to build the coarsening. We carefully adjust these matrices in order to favour clusters that meet as many constraints as possible. We show that the soft-constrained problem can be rewritten as a quadratic problem and that, in the case of \cite{mullner2011modern}, it can be solved in closed-form.

Our second contribution lies in the automation of the dendrogram (or ultrametric) interpretation. To do so, we propose a method that computes the optimal cuts of the dendrogram given the input structural constraints. The idea is to find, for each hierarchical level, the cut that minimizes the distance to the equivalent cut in the constraints tree. We prove that the resulting problem is convex and can be solved in closed-form. 

We run experiments on the Hotel dataset \cite{ModiReviews} to demonstrate our methods and compare them to their unconstrained counterparts. The dataset consists of a collection of hotel reviews, and our goal is to run hierarchical clustering algorithms on the vocabulary of the reviews in order to extract topics and lexical fields in a hierarchical fashion. Extra information is provided in the form of hierarchical must-links and cannot-links. In that case, the addition of constraints led to up to a 67\% improvement in terms of constraints compliance, and up to a 19\% improvement in terms of Dasgupta's cost~\cite{DasgCost} (a generally adopted metric in hierarchical clustering). These results support our algorithms.

%\subsection{Notations}

%Methods presented in this paper are based on graph analysis and we briefly introduce here the notations and matrices used.

%Let $\mathcal{G} = (V, E, \mathcal{W})$ be a graph, where $V$ is the set of nodes of size $N$, $E$ is the set of edges and 
%\begin{align*}
%  \mathcal{W} \colon E &\to \reals\\
%  e_{i,j} &\mapsto w_{i,j}.
%\end{align*}
%For $i,j \in \{1,..,N\} \mbox{, } e_{i,j}$ is the edge between nodes $n_i$ and $n_j$ and $w_{i,j}$ is the associated weight. Note that $w_{i,j}=0$ when there is no edge between $n_i$ and $n_j$. In our case, weights are given by the distance between the nodes points. From there, we define the distance matrix $\mathbf{D} \in \reals^{N \times N}$ by $\forall i,j \in \{1,..,N\} \mbox{, } \mathbf{D}_{i,j}=w_{i,j}$ and the laplacian matrix $\mathbf{L} = - \mathbf{D} + \Delta$, $\Delta$ is a diagonal matrix and for $i \in \{1,..,N\} \mbox{, } \Delta_{i,i} = \sum_{j=1}^N D_{i,j}$. Note that these matrices are symmetric for undirected graphs.

%\vspace{10pt}

%{\bf Organization.} The paper is organized as follows. Section 2 formalizes the general objective and introduces the optimization problem used to add soft-constraints after briefly presenting and comparing the two clustering methods. In section 3, we explain the problem of optimal hierarchical cuts and solve it mathematically. We also provide pseudo-code for our method. Finally, we test and compare our algorithms to their unconstrained counterparts on real-life datasets.

\section{Step I : Sequential Graph Coarsening}

\subsection{Problem description}

Consider a set $\mathcal{S}$ of $n$ data points. The objective is to extract clusters of points sharing similar features from $\mathcal{S}$ in a hierarchical fashion. In addition, layer-based constraints are provided by the user to add prior information on the expected clusters. In particular, the user provides a wished number of layers $\ell$, and a constraint set $\mathcal{C}_j$, $j = 1, \ldots, \ell$, for each of them. Constraints are pairwise must and cannot-links between points that can be either horizontal (when the link relates to one or two clusters from the same hierarchical level) or vertical (to allow links between clusters of different granularity levels). Horizontal constraints are specified in $\mathcal{C}_j$, while vertical constraints are precedence constraints and can be specified as cannot-link in $\mathcal{C}_j$ and subsequently as a must-link in $\mathcal{C}_{j+1}$.  

We formalized the problem of hierarchical clustering with layer-based constraints by a two-steps approach, and this section is dedicated to the first stage, which is the construction of a dendrogram based on the data space $\mathcal{S}$ and user's prior knowledge.

Consider a graph representation of the dataset $\mathcal{S}$ as $\mathcal{G} = (V, E, \mathcal{W})$, where each node in the vertex set $V$ is data point and the edges' weights $\mathcal{W}(E)$ correspond to the distance (or similarity) between the points. In this article, we will only consider the distance matrix $\mathbf{D}$ and its associated Laplacian $\mathbf{L}$, but all the results hold in the in the similarity framework. Our hierarchical clustering problem consists in sequentially merging graph nodes in a \textit{suitable} order until there is only one left. We adapted two existing methods by adding soft structural constraints. The first one, that we will call local-variation coarsening \cite{loukas2019graph} is based on the graph Laplacian $\mathbf{L}$ and chooses the reduction that minimizes spectral variations of $\mathbf{L}$. The second one is the traditional bottom-up hierarchical clustering \cite{mullner2011modern} that sequentially merges the closest nodes and computes the new distance matrix based on a predefined linkage method. The main difference between both algorithms lies on the methodology used to pick nodes to merge and compute the new distance matrix. While bottom-up linkage merely groups the two closest nodes at each step, the local-variation coarsening simultaneously merges as many pairs as possible using an edge or neighborhood-based method.

{\bf Local-variation coarsening.} We focus here on edge-based methods. Given an initial Laplacian $\mathbf{L}$ consisting of $N$ vertices, the result is a new Laplacian consisting of $\lceil N/2 \rceil$ vertices. Contraction is based on minimizing the spectral variation of the Laplacian before and after the contraction. The complexity of this algorithm is $\mathcal{O}(N\log{}N)$. The fact that the graph size is being halved at each iteration makes the method computationally lightweight. However, this might be too coarse-grained for specific datasets. 

{\bf Bottom-up approach.} This approach relies on linkage methods and the computation of the distance matrices. There are several ways to compute the new distance matrix after one contraction. Distance can be computed using one of the seven linkage methods (e.g., single, average). Naive implementations of these methods can have high complexities $(\mathcal{O}(N^3))$. However, using nearest-neighbour chains \cite{murtmult1985}, a computational complexity of $(\mathcal{O}(N^2))$ can be reached.

\subsection{Soft-constrained coarsening}

In both methods, nodes are merged by pairs using either the local variation method or the linkage clustering. These contractions are fully determined by the Laplacian or the distance matrix hence the necessity of updating them by taking constraints into account. In particular, given the coarsened Laplacian operator $\mathbf{L}_{\mathrm{c}}\in\reals^{m\times m}$, where $m<N$, coming from one coarsening step of one of the two methods, our algorithm modifies it to satisfy as many constraints as possible. The latter is achieved by solving the problem,
\begin{multline}\label{eq.qp}
\mathsf{P}(\mathbf{L}_{\mathrm{c}}) :   \min_{\mathbf{L}\in \mathcal{L} \subset \reals^{m\times m}} \, \frac{1}{2} \|\mathbf{L}-  \mathbf{L}_{\mathrm{c}}\|^2 + \\  +\frac{\lambda_1}{2} \sum_{(i,j) \in \mathrm{ML}} \|\mathbf{L}_{i,j}\|^2 +\frac{  \lambda_2}{2} \sum_{(i,j) \in \mathrm{CL}} \|\mathbf{L}_{i,j}+1\|^2,
\end{multline}
where the set $\mathrm{ML}$ represents the must links, and $\mathrm{CL}$ the cannot links, $\lambda_1, \lambda_2$ are two positive scalar weights, $\mathcal{L}$ is the set of Laplacian matrices (symmetric, with rows summing to zero and negative elements except on the diagonal), and $\|\cdot\|$ is the Frobenius norm.  

Each layer-based constraint set $\mathcal{C}_j$ contains constraints in the form of constraints matrices $\mathrm{ML}_j$ and $\mathrm{CL}_j$ whose entries are $1$ if there is a must-link or cannot-link, respectively, between datapoints, and zero otherwise. 

Problem $\mathsf{P}(\mathbf{L}_{\mathrm{c}})$ is a convex quadratic program, which can be solved efficiently by off-the-shelf solvers. An interesting feature of the bottom-up approach is that the constraint $\mathbf{L}\in \mathcal{L}$ is less important in practice and can be substituted with $\mathbf{L} \in [-1,0]^{m \times m}$. In fact, one can focus on the upper triangle of the Laplacian $\mathbf{L}_{\mathrm{c}}$ alone, and reconstruct the Laplacian a posteriori, if needed. 
This practical simplification, sensible only for the bottom-up approach, renders $\mathsf{P}(\mathbf{L}_{\mathrm{c}})$ solvable in closed-form, since each entry is now independent, as formalized next. 

\begin{proposition}[Practical simplification for the bottom-up approach]\label{prop2}
    Consider Problem $\mathsf{P}(\mathbf{L}_{\mathrm{c}})$, replacing the constraint $\mathbf{L}\in \mathcal{L}$ with $\mathbf{L} \in [-1,0]^{m \times m} $. Define $\lambda_{1,ij} =1$ if $(i,j)\in$ ML, and $0$ otherwise, and do the same for  $\lambda_{2,ij} =1$ for the cannot links. This new problem can be solved in closed-form as,
    $$
    \mathbf{L}_{ij}^* = \min\left\{\max\left\{\frac{\mathbf{L}_{\mathrm{c},ij}  - \lambda_{2,ij}}{1 + \lambda_{1,ij} + \lambda_{2,ij}}, -1 \right\}, 0 \right\}, \quad \forall j>i. 
    $$
\end{proposition}

\begin{proof} By optimality conditions.
\end{proof}

The first step of our algorithm consists in iteratively solving $\mathsf{P}(\mathbf{L}_{\mathrm{c}})$ or its simplified version, and doing a coarsening pass, multiple times till most of the constraints in $\mathcal{C}_j$ for layer $j$ are satisfied and we cannot reduce the graph anymore. Then, we continue with a new layer set $\mathcal{C}_{j+1}$, and so on. 

The final output is a large dendrogram where points connected by must-links are likely merged at a low heights and those connected by cannot-links are merged at a later layer. From there, our aim is to deduce hierarchical flat clusters without parsing the whole tree by hand. In the next section, we introduce a systematic approach to obtain such clusters given a dendrogram and layer-based constraints.

\section{Step II: Optimal Hierarchical Cuts}

\begin{figure*}
\centering
\includegraphics[width=0.8\textwidth]{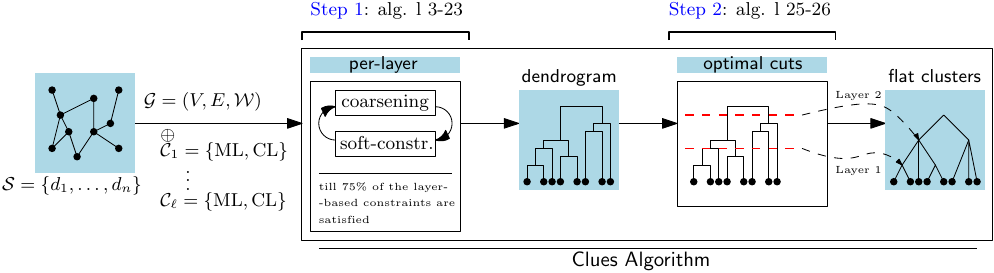}
\caption{\footnotesize Schematic diagram of \clues' main steps.}
\label{fig.setup}
\end{figure*}

\subsection{Problem Description}

Given a dendrogram, the problem we want to solve is how to obtain hierarchical flat clusters satisfying the horizontal and vertical constraints. The usual approach would be to parse the dendrogram manually, but this is hardly an option for large-scale trees. Another possibility is to cluster from the layers that we have built. But this may be sub-optimal, since not all the constraints may have been satisfied. Here we propose a different approach. 

Consider $\ell$ layers. For each layer $j \in {1,2,\ldots, \ell}$, we consider the clusters that Step I has produced. These are indicated with $c_{i,j}, i = 1, \ldots, n_i$ as for cluster $i$, in layer $j$. As we are parsing a dendrogram, which is a binary tree, we can order the branches from the data at height $0$ to the root at height $H>0$. We can then define two heights for each $c_{i,j}$. 

The first is $h_{i,j}$ as the minimal height at which all the must-link of $c_{i,j}$ are satisfied. If Step I is completely successful, the height $h_{i,j}$ will correspond to the height derived from Step I, otherwise it will be higher. By construction, we know that such height exists, since eventually all the data will be merged in a single point at $H$. 
The second height is $H_{i,j}$, defined as the minimal height greater or equal than $h_{i,j}$ at which one cannot-link of $c_{i,j}$ is violated. Once again, if Step I is successful $H_{i,j}$ could be equivalent to $h_{i,j}$ or strictly higher. If $c_{i,j}$ has no cannot link, then $H_{i,j} = H$. 

Then, we define the optimal level cut as the problem of finding the height at which to cut the dendrogram for each level $j$, such that we minimize constraint violation,
\begin{equation}
\tilde{h}_j \in \mbox{argmin}_{h \in \reals_{+}} \sum_{i=1}^{n_i} \mathcal{D}_{i,j}(h),
\label{optcut}
\end{equation}
where $\mathcal{D}_{i,j}: \reals_{+} \to \reals_{+}$ is the function defined as the distance between $h$ and its convex projection onto the interval $[h_{i,j}, H_{i,j}]$, that is $\mathcal{D}_{i,j}(h) = |h-\max\{\min\{h, H_{i,j}\}, h_{i,j}\}|$. 

\begin{proposition}\label{oc}
Let $\hat{h}_1 \leq \hat{h}_2 \leq \ldots \leq \hat{h}_{2n_i}$ be the elements of $\{h_{1,j}, H_{1,j},\ldots, h_{n_i,j}, H_{n_i,j}\}$ sorted in increasing order. Problem \ref{optcut} is convex and any point of the set $\{h | h \in [\hat{h}_{n_i}, \hat{h}_{n_i+1}]\}$ is a solution for \eqref{optcut}.
\end{proposition}

\begin{proof}{(Sketch)}
Convexity of \eqref{optcut} follows from the convexity of $\mathcal{D}_{i,j}$, which can be proved by construction. Then, writing the optimality conditions, the solution set can be derived. 
\end{proof}

As in Step I, Step II is based on a problem that can be solved in closed-form and its resolution is computationally lightweight. We are now ready to discuss the overall algorithm.

\subsection{Overall algorithm}

We now present our main algorithm, labeled \clues, as to indicate that we are \emph{clu}stering with soften prior information (that is: clues), we describe it in Algorithm~\ref{alg:clues}, and it performs hierarchical flat clusters extraction using the soft-constrained Problem~\ref{eq.qp} and the optimal cuts given by Prop.~\ref{oc}. \clues\ is also depicted in Figure~\ref{fig.setup}.

\begin{algorithm}%[H]
    \caption{\clues }\label{alg:clues}
    \small
    \begin{algorithmic}[1]
        \Require data set $\mathcal{S}$ of $n$ data points, distance matrix $\mathbf{D}$, smoothing parameters $\lambda_1$, $\lambda_2$, level-based pairwise constraints $\mathcal{C}_j$, coarsening method, $I_{\max}>0$
        \Ensure Hierarchical flat clusters
        \State Build the graph $\mathcal{G}$
        \State{{\color{blue} // Step I}}
        \For{Layers $k = 1, \ldots, j$} 
        \State $I = 0$
        \If{method is traditional bottom up clustering}
        %\Comment{{\color{blue} method is local variation coarsening}}
        \State Update $\mathbf{D}$ using Prop.~\ref{prop2}
            \While{$75\%$ of $\mathcal{C}_j$ is not satisfied and $I<I_{\max}$}
            \State Merge the two closest nodes
            \State Compute the new distance matrix $\mathbf{D}$
            \State Update the new $\mathbf{D}$ using Prop.~\ref{prop2}
            \State Update $\mathcal{G}$, $I \leftarrow I+1$
            %\State $I \leftarrow I+1$
            \EndWhile
        \Else
        \Comment{{\color{blue} method is local variation coarsening}}
        \State Build the Laplacian $\mathbf{L}$
        \State Update $\mathbf{L}$ using Problem \ref{eq.qp}
        \While{$75\%$ of $\mathcal{C}_j$ is not satisfied and $I<I_{\max}$}
        \State Merge as many pairs of nodes as possible
        \State Compute the new Laplacian $\mathbf{L}$
        \State Update $\mathbf{L}$ using Problem \ref{eq.qp}
        \State Update $\mathcal{G}$, $I \leftarrow I+1$
        %\State $I \leftarrow I+1$
        \EndWhile
        \EndIf
        \EndFor
        \State{{\color{blue} // Step II}}
        \State Compute the set of optimal cuts $C$ of the dendrogram $\mathcal{D}$ resulting from the clustering using Prop.~\ref{oc}
        \State Cut $\mathcal{D}$ using $C$ and derive the hierarchical flat clusters
    \end{algorithmic}
\end{algorithm}

If the method is the traditional bottom-up clustering, the algorithm works by locally looking at the distance matrix values and treating edges independently. Hence, using the closed-form alternative given in Prop.~\ref{prop2} leads to satisfying results and helps keeping control of the complexity (which is $\mathcal{O}(n^3)$, and $\mathcal{O}(n^2)$ when using nearest-neighbour chains).

As for the local variation clustering, we solve Problem~\ref{eq.qp} at each step to exploit the neighborhood information contained in the diagonal term. The worst-case complexity is $\mathcal{O}(n^3)$. 

\section{Results}

{\bf A first example.} We test our algorithm on an hotel reviews database~\cite{ModiReviews}, with the aim of comparing themes and lexical fields across the different states in the USA. The data consist of $n$ reviews and meta-data containing, e.g., timestamps and locations. We use the former to build our vocabulary and dendrogram and the latter to locate the hotel. 

\begin{figure}
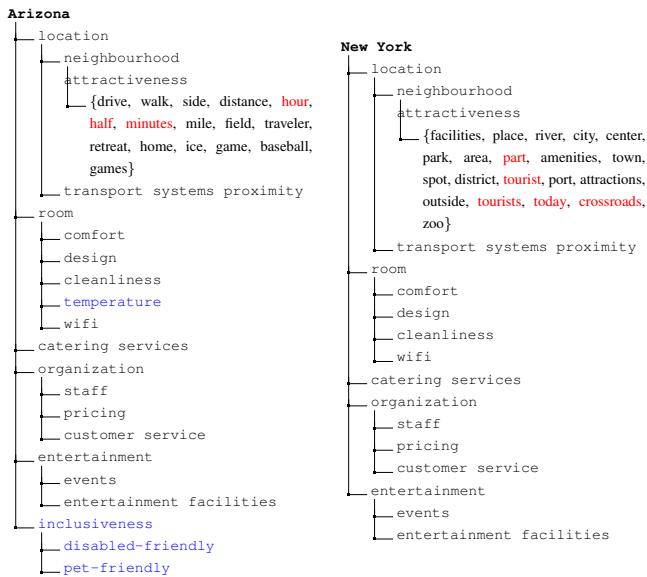

\centering
\scalebox{0.7}{%
\begin{minipage}[c]{0.7\linewidth}
\footnotesize
\dirtree{%
.1 \textbf{Arizona}.
.2 location.
.3 neighbourhood attractiveness.
.4 \rm\{drive, walk, side, distance, \textcolor{red}{hour}, \textcolor{red}{half}, \textcolor{red}{minutes}, mile, field, traveler, retreat, home, ice, game, baseball, games\}.
.3 transport systems proximity.
.2 room.
.3 comfort.
.3 design.
.3 cleanliness.
.3 \textcolor{blue}{temperature}.
.3 wifi.
.2 catering services.
.2 organization.
.3 staff.
.3 pricing.
.3 customer service.
.2 entertainment. 
.3 events.
.3 entertainment facilities.
.2 \textcolor{blue}{inclusiveness}. 
.3 \textcolor{blue}{disabled-friendly}.
.3 \textcolor{blue}{pet-friendly}.
}
\end{minipage} \hfill
\begin{minipage}[c]{0.7\linewidth}
\footnotesize
\dirtree{%
.1 \textbf{New York}.
.2 location.
.3 neighbourhood attractiveness.
.4 \rm\{facilities, place, river, city, center, park, area, \textcolor{red}{part}, amenities, town, spot, district, \textcolor{red}{tourist}, port, attractions, outside, \textcolor{red}{tourists}, \textcolor{red}{today}, \textcolor{red}{crossroads}, zoo\}.
.3 transport systems proximity.
.2 room.
.3 comfort.
.3 design.
.3 cleanliness.
.3 wifi.
.2 catering services.
.2 organization.
.3 staff.
.3 pricing.
.3 customer service.
.2 entertainment. 
.3 events.
.3 entertainment facilities.
}
\end{minipage}
}
\caption{\footnotesize Hierarchical topics extraction running Clues on hotel reviews. Topics names have been inferred manually from the lexical fields. An example is given for the sub-cluster ``attractiveness" for comparative purposes.}
\vskip-0.5cm
\label{fig.example}
\end{figure}

Example trees of hierarchical topic extraction for New York and Arizona reviews are displayed in Figure~\ref{fig.example}, setting two layers. Clusters labels were built manually from the output lexical fields. At this granularity level, clusters are similar for both states as they are guided by the constraints. Nonetheless, we can already notice a novel theme (inclusiveness) and sub-theme (temperature) in the Arizona tree that were not defined in the constraints (indicated in blue on the trees). Thus, using constraints to give a hierarchical structure that guides the algorithm helps achieving consistent results while still allowing the extraction of novel information.

The main source of information lies in the vocabulary associated with each cluster. Labels might be subjective or poorly set, hence the necessity to check results at the finest level. For instance, even though labels are the same, the lexical fields differ according the the state. In Figure~\ref{fig.example}, we have also displayed the lexical fields behind ``neighbourhood attractiveness". While both related to the interests of the hotel area, the lexical fields are different and illustrate the heterogeneity in tourists' expectations according to the area. Note that some words in the lexical fields (in red) are inconsistent or meaningless. These words constitute errors of the algorithm.

%Of course, the number of hierarchical levels is an argument of the algorithm and can be adapted to fit with the user's expectations. After one pass of the algorithm, sub-clusters might still be too coarse and one may want to add another finer layer to describe them.

%Parameters $\lambda_1$ and $\lambda_2$ in \eqref{eq.qp} (as well as their ratio $\frac{\lambda_1}{\lambda_2}$) are extremely sensitive and have a great impact on the results. 
%Not only they influence the building of the dendrogram by setting the strength of the constraints, but they also widely impact the optimal cut heights for the same reasons. In our configuration, we found out that $\lambda_1 \mbox{, } \lambda_2 \in [0,0.1] \mbox{ and } \frac{\lambda_1}{\lambda_2} \approx 1$ lead to more consistent and balanced clusters. A high $\lambda_1$ will give a lot of importance to must-links; they will all be merged during the very first steps of the clustering process leading to a very low first cut height. 
%On the other hand, an excessively large $\lambda_2$ will lessen or even cancel the impact of must-links hence optimal cuts computation almost only take cannot-links into consideration leading to inconsistent final clusters.
%\footnote{Even with reasonable values for $\lambda_1$ and $\lambda_2$, clustering quality is highly impacted by very small variations in the parameters. Moreover, we noticed that ``ideal" parameters values vary quite a lot depending on the initial database.}

%\subsection{Discussion}

{\bf Comparisons.} We move to assess the performances of \clues\ based on three different criteria: \emph{(i)} the widely used Dasgupta's cost~\cite{DasgCost} that measures the resulting dendrogram ``quality'' (the lower the cost the better); \emph{(ii)} the percentage of violated constraints; and \emph{(iii)} the runtime of the algorithm. We also compare \clues\ to its unconstrained version, where we do not perform the soft-constrained coarsening in Step I, i.e., we use the algorithms of \cite{loukas2019graph} and \cite{mullner2011modern} as they are.

\begin{figure}
      \centering
      \resizebox{0.45\textwidth}{!}{\input{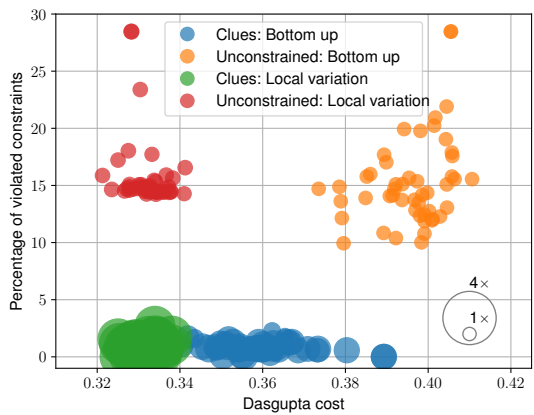}}
      \caption{\footnotesize Performance of \clues\ and comparison: lower Dasgupta costs are associated with higher quality hierarchical clustering hence points on the lower left represent the best results. The sizes of the points are proportional to the normalized time taken to run \clues\ on the corresponding dataset.}
      \label{fig.cr}
\end{figure}

As depicted in Figure~\ref{fig.cr}, where we use 50 different review datasets, \clues\ obtains the best results for constraints enforcement (leading to up to a 67\% improvement, and a below $5\%$ overall constraint violation), and it can further reduce the Dasgupta's cost in the bottom-up approach (up to a 19\% improvement). Time-wise, we use the size of the dots to represent the normalized run-time, and we can see that the local-variation version of \clues\ is the most demanding. This is due to the use of CVX to solve the soft-constrained problem~\eqref{eq.qp}, and it represents a $\sim 3 \times$ increase in computational time. 

The 50 datasets are extracted from the hotel reviews sorted by state. This enables to have sets of different sizes (from $\sim 25$ to $\sim 1250$ points) and varied vocabulary while keeping the same constraints.

\section{Conclusions}

In this paper, we introduced an approach to perform semi-supervised hierarchical clustering on large databases. The specificity of our method being the possibility to extract flat clusters automatically from word data by exploiting structural and user-defined constraints, making it easier to analyze the results. Constraints have to be provided in the form of layer-based must-links and cannot-links and are used throughout the whole process to guide the clustering and extract the final flat non-binary tree.
Performances of our method (named \clues) are evaluated on a benchmark dataset and the results encourage further work in this direction. %Among others, it has been noticed that the amount and quality of constraints played a significant role on the algorithms' performances.

%Possible improvements of our method include automatic definition of labels using the emerging artificial intelligence tools. This might make the coarse tree more self-explanatory due to specific labels and thus spot main differences between trees at first glance. Another upgrade would be to fine-tune the optimization solver for local-variation based methods in order to reduce the time complexity of the algorithm.

%describe database and practical applications that suits the described method
%discuss results and efficiency

\bibliographystyle{ieeetr}
\bibliography{bibliography}

\end{document}